\documentclass{article}

\usepackage[dblblindworkshop,final]{neurips_2026}
\workshoptitle{The Third Workshop on GenAI for Health: Agentic Systems, Clinical Trust, and Future Potential (GenAI4Health)}

\usepackage[utf8]{inputenc}
\usepackage[T1]{fontenc}
\usepackage[hidelinks]{hyperref}
\usepackage{url}
\usepackage{booktabs}
\usepackage{amsmath, amssymb, amsfonts}
\usepackage{microtype}
\usepackage{graphicx}
\usepackage{tabularx}
\usepackage{makecell}
\usepackage{float}
\usepackage{enumitem}
\usepackage{xcolor}
\usepackage[capitalise,nameinlink]{cleveref}

\graphicspath{{figures/}}

\crefname{appendix}{appendix}{appendices}
\Crefname{appendix}{Appendix}{Appendices}

\newcommand{\thd}[1]{{\bfseries #1}}
\definecolor{curaRed}{HTML}{C0392B}
\definecolor{curaGreen}{HTML}{2E8B57}
\setlist[itemize]{leftmargin=*,itemsep=0.25em,parsep=0em,topsep=0.25em}

\title{Cura 1T: Healthcare Foundation Model via Recursive Self-Improvement}

\author{%
  Haolin Chen, Leon Qi, Steve Brown, Deon Metelski \\
  \textbf{Tao Xia, Joonyul Lee, Qixuan Wang, Kevin Riley} \\
  \textbf{Frank Wang, Weiran Yao} \\
  \normalfont Actava AI
}

\begin{document}

\maketitle

\begin{abstract}
Healthcare spans high-stakes communication, expert reasoning, and workflow execution, yet
specialized language models that cover these use cases together remain limited. A healthcare model
must handle patient consultation, clinical reasoning over text and images, interactive diagnosis,
and electronic health record (EHR) tool use. These capabilities fail in different ways, and a narrow
update for one task can degrade another. We present Cura 1T, a healthcare foundation model trained through
recursive self-improvement (RSI). In each RSI round, the RSI harness runs the current model on healthcare
benchmarks, evaluates the trajectories to locate capability gaps, and refines the training mixture by synthesizing training data.
On 6 healthcare benchmarks, Cura 1T scores highest on MedAgentBench, HealthBench Professional, HealthBench Hard, MedXpertQA
text, and AgentClinic, and second on MedXpertQA multimodal. It preserves performances on out-of-domain reasoning and agentic benchmarks including
AIME, GPQA-Diamond, and $\tau^2$-Bench.
\end{abstract}

\section{Introduction}
\label{sec:intro}

Frontier language models now answer difficult clinical questions~\citep{medpalm}, but healthcare
deployment needs a wider range of skills and tolerates fewer mistakes. We focus on three use
cases. \emph{Patient care} requires clinician- and patient-facing
responses that follow physician guidelines~\citep{medpalm,healthbench,healthbenchpro}.
\emph{Clinical reasoning} requires solving expert medical questions across text and
images~\citep{medqa,medmcqa,medxpertqa}. \emph{Healthcare agentic tasks} require multi-turn
diagnostic workups, electronic-health-record (EHR) tool use, and long-horizon workflows under strict
task protocols~\citep{agentclinic,medagentbench,chibench,physicianbench,healthagentbench,fhir}.
These tasks are related, but they fail in different ways: rubric omissions, missing facts, brittle
reasoning, premature diagnoses, and malformed tool calls -- each call for different training signals. Each
piece has advanced separately: expert QA~\citep{medqa,medmcqa,medxpertqa}, rubric-graded
clinical conversations~\citep{healthbench,healthbenchpro}, diagnosis and EHR-tool
benchmarks~\citep{agentclinic,medagentbench,chibench,physicianbench,healthagentbench}, and
continuous-care systems~\citep{baichuanm4}. Few specialized models cover all three together.

Building such a model is a data-construction problem as much as a model-training problem. Compared
with coding or mathematics, healthcare has less abundant training signal, and its sensitivity makes
high-quality supervision harder to obtain. Useful examples are fragmented across guidelines, exams,
patient interactions, images, and EHR workflows, and many outcomes cannot be graded by a simple
executor or answer key. Because failure modes are unevenly distributed across task categories,
adding examples for one behavior can improve that behavior while eroding behavior that was already
correct elsewhere. The training algorithm and the hyperparameters are not the bottleneck. The work
is to identify the missing data in the recipe, add publicly available or
synthetic surrogates, and curate mixtures that transfer across tasks without unnecessary
forgetting.

We build Cura 1T, a healthcare foundation model post-trained on the one-trillion-parameter Kimi-K2.6
base~\citep{kimik26} through recursive self-improvement. Running the full post-training cycle by
hand (planning a run, launching training, evaluating the result, reading trajectories, diagnosing
failures, and rebuilding the mixture) costs substantial expert time and slows iteration; recent work
delegates comparable cycles to agents~\citep{autoresearchrepo,aiscientistv2,alphaevolve}. In each
round, the agent evaluates the current model on the relevant benchmarks, reads the graded
trajectories to locate a capability gap, sets the target behavior and acceptance criteria, and
synthesizes training data with specialized data-construction skills. \Cref{sec:loop} describes how candidates are scored and accepted or
reverted.

\Cref{tab:headline} summarizes the main result: the same consolidated model improves over its base
across patient-care, clinical-reasoning, and healthcare-agentic evaluations. Against frontier
comparators, Cura 1T is the strongest model on five of the six healthcare benchmarks and ranks second
on MedXpertQA multimodal (\Cref{sec:eval}). On out-of-domain benchmarks in mathematics,
scientific reasoning, and general agentic tasks, Cura 1T holds the base model's performance (See \Cref{fig:ood-comparison}).

Our contributions are: (i) Cura 1T, a single model covering patient care, clinical reasoning over
text and images, and healthcare agentic tasks, which scores highest on five of the six healthcare
benchmarks and holds its base model's out-of-domain performance; (ii) the recursive self-improvement
loop that produced it, whose search object is the post-training data mixture rather than prompts,
code, or hyperparameters, reported round by round including the reverted candidates; and (iii)
reusable data-construction skills---reasoning correction, knowledge injection, behavior calibration,
and retention anchoring---that map graded failure modes to targeted synthetic data.

\begin{table}[h]
  \caption{Improvement of Cura 1T over its base model. Metrics differ by benchmark (rubric score for
  HealthBench, exact-letter pass@1 for MedXpertQA, task success for AgentClinic and MedAgentBench).}
  \label{tab:headline}
  \centering
  \footnotesize
  \renewcommand{\arraystretch}{1.12}
  \setlength{\tabcolsep}{6pt}
  \begin{tabular*}{0.72\linewidth}{@{\extracolsep{\fill}}lrrr@{}}
    \toprule
    \thd{Benchmark} & \thd{Base} & \thd{Cura 1T} & \thd{$\Delta$} \\
    \midrule
    MedAgentBench            & 0.847 & \textbf{0.940} & $+0.093$ \\
    HealthBench Professional & 0.503 & \textbf{0.662} & $+0.159$ \\
    HealthBench Hard         & 0.222 & \textbf{0.368} & $+0.146$ \\
    MedXpertQA               & 0.569 & \textbf{0.655} & $+0.086$ \\
    AgentClinic              & 0.754 & \textbf{0.796} & $+0.042$ \\
    \bottomrule
  \end{tabular*}
\end{table}

\section{Related Work}
\label{sec:related}

\paragraph{Healthcare AI.}
Med-PaLM showed that large models can encode enough medical knowledge to answer licensing-exam
questions~\citep{medpalm}, and the Baichuan series, in particular Baichuan-M4, extends medical
modeling toward continuous care with tool use, patient memory, action constraints, evidence
retrieval, and multimodal perception~\citep{baichuanm4}. Benchmarks broadened too.
HealthBench grades open-ended clinical responses against physician-written rubrics~\citep{healthbench},
and HealthBench Professional extends this setting to real clinician chats~\citep{healthbenchpro}.
Clinical-reasoning benchmarks began with medical-exam question answering~\citep{medqa,medmcqa};
MedXpertQA increases difficulty and adds multimodal cases with real clinical images and
context~\citep{medxpertqa}. Agentic benchmarks evaluate whether models can act over time:
AgentClinic tests multi-turn diagnosis in a simulated clinic~\citep{agentclinic}, MedAgentBench
evaluates clinically derived EHR tasks against a standards-compliant record~\citep{medagentbench,fhir},
and CHI-Bench, PhysicianBench, and HealthAgentBench extend evaluation toward long-horizon,
role-composed, policy-driven healthcare workflows~\citep{chibench,physicianbench,healthagentbench}.
We use these benchmark families to evaluate Cura 1T across the three use cases of \Cref{sec:intro}.

\paragraph{Post-training.}
Post-training commonly combines supervised adaptation, reinforcement learning, self-training, and
parameter-efficient updates. Rejection sampling fine-tuning trains on a model's own highest-scoring
generations~\citep{rft,llama2}, and STaR bootstraps reasoning by retaining sampled rationales that
reach the correct answer~\citep{star}. Self-distillation fine-tuning (SDFT) uses the model's own
samples as the student distribution and a teacher conditioned on additional context as the target,
keeping updates close to the base model's generation policy~\citep{shenfeld2026sdft}. These methods
are often combined with reasoning-aware data formats that preserve chain-of-thought
behavior~\citep{cot,deepseekr1}.

\paragraph{Recursive self-improvement and automated research.}
Self-Refine and Reflexion use natural-language critique or task feedback to improve subsequent
attempts, while OPRO, DSPy, and TextGrad treat prompts, programs, or pipeline components as objects
to optimize against a metric~\citep{selfrefine,reflexion,opro,dspy,textgrad}. The public autoresearch
repository applies this to a coding agent that edits a real training loop, runs fixed-budget
experiments, and keeps or discards each change~\citep{autoresearchrepo}. AI Scientist,
AI Scientist-v2, and AlphaEvolve extend the pattern to scientific workflows and algorithm discovery,
while ResearchGym and recent surveys note that reliability, provenance, and reproducibility remain
central bottlenecks~\citep{aiscientist,aiscientistv2,alphaevolve,researchgym,autoresearchai}. Cura
keeps the closed loop and changes what is searched. Instead of prompts, code, or hyperparameters, the
loop curates the post-training data mixture that is then trained into the model, and each accepted
round supplies the starting point for the next.

\section{Cura 1T}
\label{sec:system}

Cura 1T is post-trained on top of Kimi-K2.6~\citep{kimik26} for patient care, clinical reasoning,
and healthcare agentic tasks. The released model uses a 256K context window and native text and
vision input. Patient care covers clinician- and patient-facing responses that must follow physician
rubrics while remaining clear and safe. Clinical reasoning covers expert medical question answering
across text and images, where the model must combine factual recall, differential reasoning, and
answer selection. Healthcare agentic tasks cover multi-turn clinical workflows, where the model must
gather evidence, use tools, and complete diagnosis or EHR tasks before producing a final answer.
Training uses a lightweight training stack (\Cref{sec:training}) driven by the recursive
self-improvement loop (\Cref{sec:loop}).

\subsection{Training}
\label{sec:training}

Supervised fine-tuning (SFT) is the low-cost prerequisite step inside each round. Before launching a
more time-consuming RL or SDFT run, the agent uses SFT to verify that the proposed data mixture and
hyperparameters are well formed, train stably, and move the target metrics in the intended
direction. Once this screen passes, the round proceeds through reinforcement learning (RL), followed
by self-distillation fine-tuning (SDFT)~\citep{shenfeld2026sdft} as the final step.

SDFT trains the model from its own on-policy samples toward a teacher distribution conditioned on
additional information. For a prompt $x$, privileged context $c$, and student sample
$y \sim \pi_\theta(\cdot\mid x)$,
\begin{equation}
  \mathcal{L}(\theta)
    = D_{\mathrm{KL}}\!\left(\pi_\theta(\cdot\mid x)\,\middle\|\,\pi(\cdot\mid x,c)\right)
    = \mathbb{E}_{y \sim \pi_\theta(\cdot\mid x)}
      \left[\log \frac{\pi_\theta(y\mid x)}{\pi(y\mid x,c)}\right],
  \label{eq:sdft}
\end{equation}
where $\pi_\theta(\cdot\mid x)$ is the student and $\pi(\cdot\mid x,c)$ the privileged-context
teacher. In Cura, $c$ is the extra information used to generate a clean teacher trajectory: an
intervened or augmented trajectory, a reference behavior, or verified knowledge used to ground
teacher generation. The student sees only the original prompt after this context is removed. The
update is therefore anchored to trajectories the model can itself produce. In long medical reasoning
traces, copying an off-policy chain of thought damages the model's native reasoning behavior.

All experiments follow the same protocol. With optional human guidance, the training agent writes a
plan, trains candidate models through the SFT-to-RL-to-SDFT stack, evaluates the model, and reports
the evidence for a keep-or-revise decision. We run the loop separately for capabilities that need
improvement. The loop is recursive in that a kept round becomes the continuation point for the next
iteration and contributes its validated data refinement to the growing mixture. A reverted round remains in the experimental
record but is not used as the basis for later improvement, because the intervention caused severe
side effects or was not favored. Cura 1T is trained from the consolidated mixture accumulated across
completed capability loops, rather than from a single benchmark-specific update. Final
hyperparameters are reported in \Cref{app:training-config}.

\begin{figure}[t]
  \centering
  \includegraphics[width=0.95\linewidth]{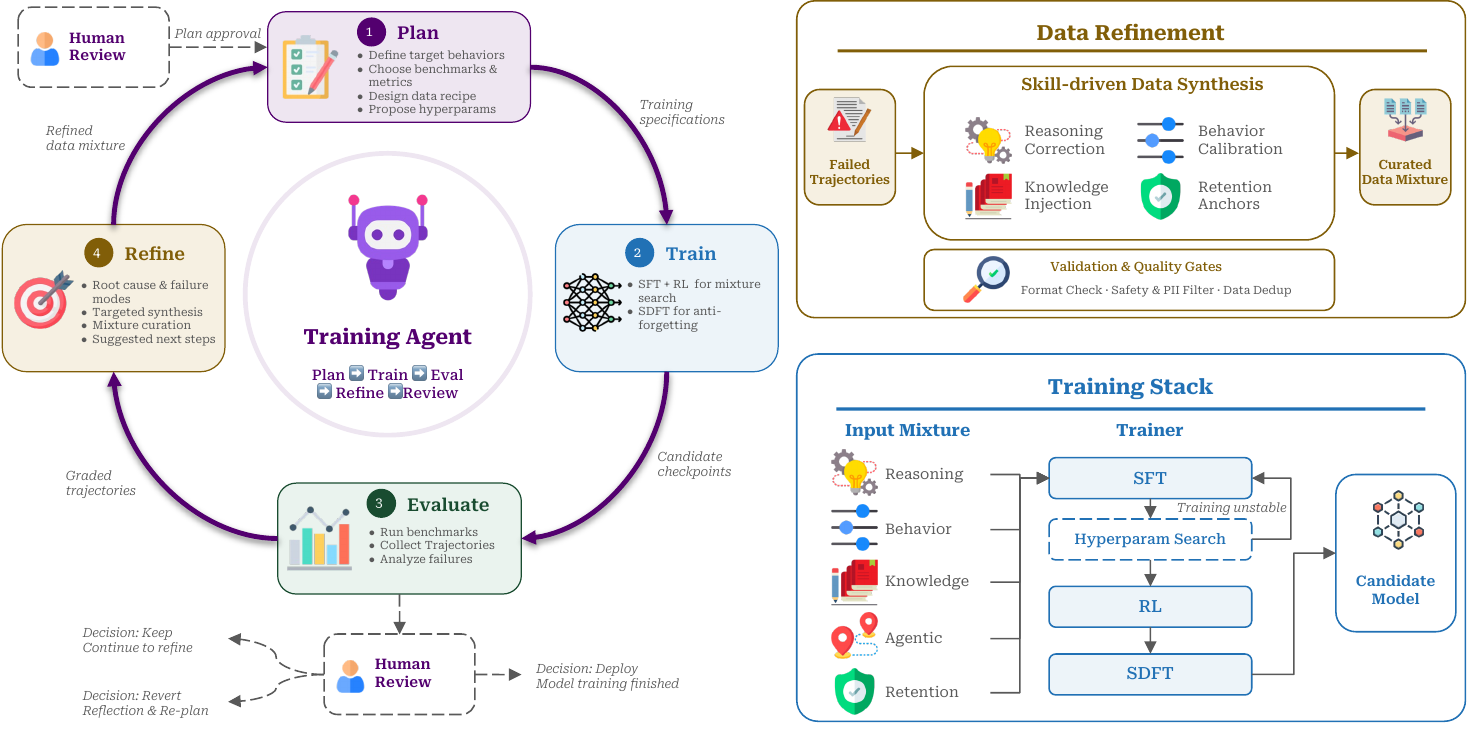}
  \caption{Left: recursive self-improvement loop for Cura 1T. A reviewer approves the plan before
  training and makes the keep, revert, or deploy decision after evaluation. Right: data refinement
  pipeline and training stack.}
  \label{fig:loop}
\end{figure}

\subsection{Recursive self-improvement loop}
\label{sec:loop}

In healthcare post-training the limiting resource is scarce and uneven domain data for the use cases
above, not a fixed benchmark dataset with a nearly finished recipe. Recent self-improvement and automated-research workflows search over hypotheses,
programs, skills, experiment trees, or training
configurations~\citep{opro,dspy,textgrad,autoresearchrepo,aiscientistv2,alphaevolve}. The agent
writes configurations and monitors training, but its central task is to analyze failures, synthesize targeted data, and
curate a mixture that preserves previously solved behavior.

\Cref{fig:loop} shows the loop. \emph{Plan} defines target behaviors,
benchmarks, metrics, data recipe, and candidate hyperparameters, then passes through human plan
approval. \emph{Train} runs model training, using SFT as the mixture and hyperparameter
screen, RL as the reward-driven improvement stage, and SDFT as the final trainer of the round.
\emph{Evaluate} runs the benchmarks, collects trajectories, and analyzes failures. \emph{Refine}
uses failed trajectories as evidence for root-cause analysis, converts the resulting failure modes
into targeted data, curates the next mixture, and validates the candidate rows. In \emph{Review}, a
person inspects both the validated data refinement and the candidate model, then issues a keep,
revert, continue-to-refine, or deploy decision before the next round begins.
\Cref{app:components} details the role and output of each component.

The agent first categorizes benchmark failures, then uses the resulting failure patterns to create
new training examples that target the same missing capability while preserving retained behavior.
The benchmark examples remain held out from training; their scores guide the keep-or-revise decision
at the end of each round.
Data synthesis is driven by the following agent skills:
\begin{itemize}
  \item \textbf{Retention anchor.} Add a small set of examples for capabilities the current model already handles, so a targeted update does not overwrite them.
  \item \textbf{Reasoning correction.} Identify the failed reasoning step, obtain a corrected explanation from a frontier model, and use the resulting pattern to synthesize new cases in different clinical contexts. The corrected explanation supplies the privileged context $c$ in \Cref{eq:sdft}.
  \item \textbf{Knowledge injection.} Identify missing clinical knowledge, ground it with retrieved documents, and synthesize new closed-book QA rows with explicit reasoning.
  \item \textbf{Behavior calibration.} Compare desired and observed answer behavior, then use a rubric, fix guidance, or a reference response to generate new cases that exercise the desired response.
  \item \textbf{Data mixture curation.} Merge synthesized correction rows with retention anchors, then deduplicate, cap, and reweight the candidate training set.
\end{itemize}
Curation changes one mixture decision at a time: add retention anchors when an update causes
forgetting, increase targeted synthetic trajectories when a failure mode remains underrepresented, or
remove rows that dilute the target distribution. The agent handles task-specific failures outside
these skills, such as agentic workflow errors, case by case.

\section{Evaluation}
\label{sec:eval}

\subsection{Setup}

We evaluate Cura 1T on healthcare benchmarks covering patient-facing response quality, clinical
reasoning, interactive diagnosis, and EHR tool use. MedAgentBench measures task success on
clinically derived FHIR tool-use tasks~\citep{medagentbench}. HealthBench Professional and
HealthBench Hard score open-ended answers with physician-authored
rubrics~\citep{healthbench,healthbenchpro}. MedXpertQA is exact-letter graded and reported as text,
multimodal, and overall pass@1~\citep{medxpertqa}. AgentClinic evaluates interactive diagnosis in a
simulated clinic; we use the tool-based harness of \Cref{app:harness} and report per subset and
overall pass@1~\citep{agentclinic}. All benchmarks are evaluated at $T{=}1.0$, except that the
MedAgentBench development path in \Cref{fig:evolution-across-benchmarks} uses $T{=}0.6$. In every
table, rows labelled Round $k$ are candidate models produced by one round, shown with the
accept or revert decision; the Cura 1T row is the released model trained from the consolidated mixture
after each evolution path saturates.

\begin{figure}[t]
  \centering
  \includegraphics[width=0.9\linewidth]{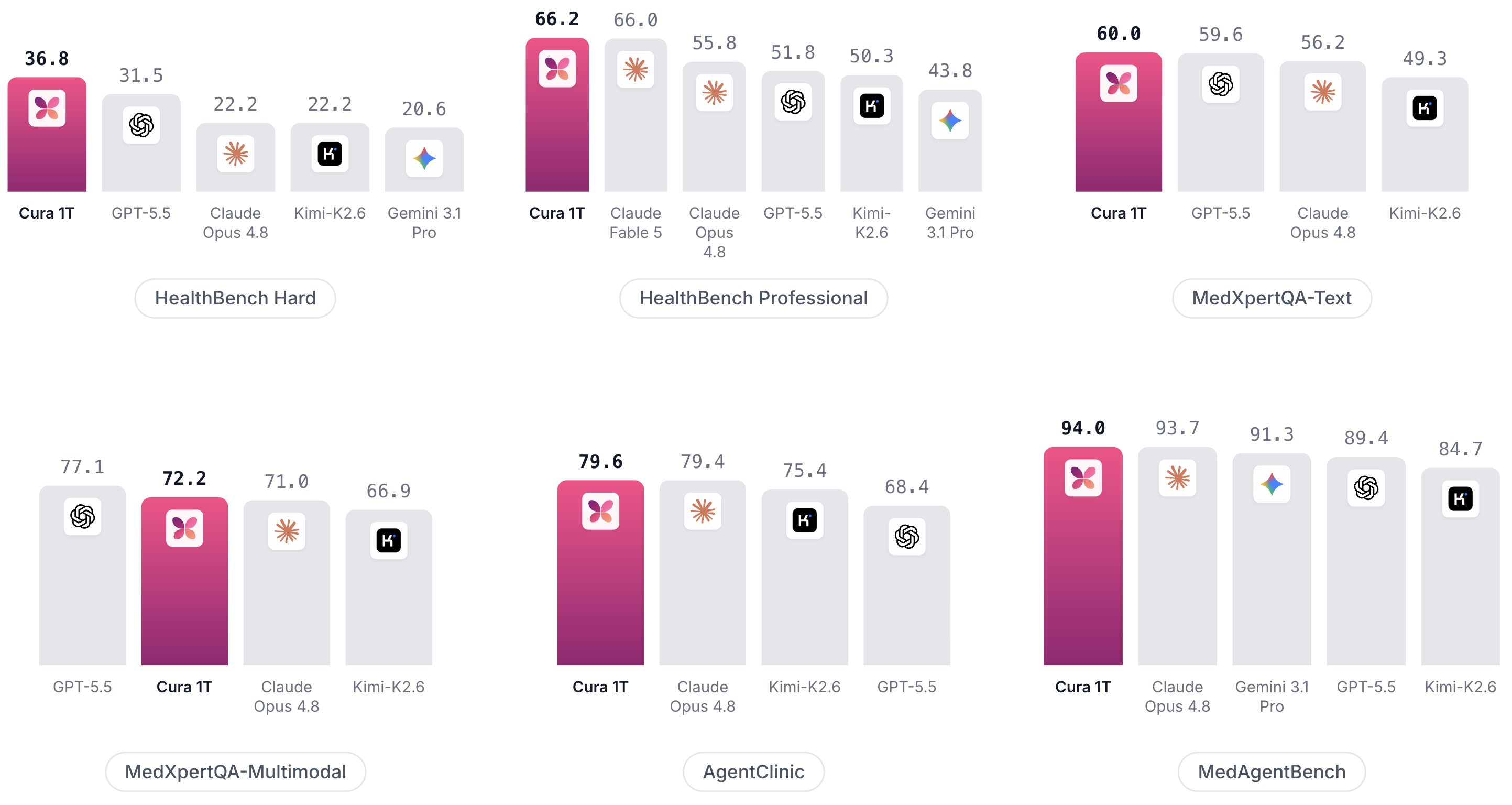}
  \caption{Cura 1T against frontier models and the Kimi-K2.6 base across six healthcare benchmarks:
  MedAgentBench~\citep{medagentbench}, HealthBench Professional and
  Hard~\citep{healthbench,healthbenchpro}, MedXpertQA text and multimodal~\citep{medxpertqa}, and
  AgentClinic~\citep{agentclinic}.}
  \label{fig:cover}
\end{figure}

\subsection{MedAgentBench}
\label{sec:mab}

MedAgentBench tests whether a model can execute clinically derived EHR workflows through FHIR tool
calls. We evaluate the model as a native tool-caller against a running FHIR server, using renderer-native function calling with read,
write, and finish actions. The principal failure mode is brittle execution of EHR writes. The base
model often identifies the intended clinical action but emits a resource that is rejected because a
required field is missing, a coded value is wrong, or the payload is inconsistent with the
request. \emph{Behavior calibration} targets these errors with synthetic trajectories that exercise
the same FHIR write patterns in new patient contexts. \emph{Behavior calibration + retention} adds successful
trajectories from surrounding tool behaviors so that a narrow update does not degrade the rest of
the workflow. \emph{Harness bug fix} corrects the evaluation configuration and restores the intended
prompt. \Cref{tab:mab} shows that the rounds raise task success from 0.847 to 0.973. The round-3
checkpoint targets this capability alone, whereas the released model folds the corresponding synthetic
examples into the consolidated mixture and has to hold its scores on the other benchmarks at the same
time, which costs 0.033 here; at 0.940 it still exceeds the strongest frontier reference at 0.937.

\begin{table}[h]
  \caption{MedAgentBench task success. Bold marks the best score and underline the second-best
  distinct score.}
  \label{tab:mab}
  \centering
  \footnotesize
  \renewcommand{\arraystretch}{1.1}
  \setlength{\tabcolsep}{5pt}
  \begin{tabular}{@{}llrl@{}}
    \toprule
    \thd{Model} & \thd{Intervention} & \thd{Overall} & \thd{Decision} \\
    \midrule
    Kimi-K2.6 & $-$ & 0.847 & $-$ \\
    Round 1 & Behavior calibration & 0.943 & keep \\
    Round 2 & Behavior calibration + retention & \underline{0.967} & keep \\
    Round 3 & Harness bug fix & \textbf{0.973} & keep \\
    Cura 1T & Consolidated data mixture & 0.940 & release \\
    \midrule
    Claude Opus 4.8 & $-$ & 0.937 & $-$ \\
    GPT-5.5 & $-$ & 0.894 & $-$ \\
    Gemini 3.1 Pro & $-$ & 0.913 & $-$ \\
    \bottomrule
  \end{tabular}
\end{table}

\subsection{HealthBench}
\label{sec:hb}

HealthBench evaluates whether a model gives clinically appropriate open-ended answers rather than
only selecting a multiple-choice option. The base model scores 0.503 on Professional and 0.222 on
Hard. Failures are dominated by omitted required points rather than explicit penalty violations, and
rejection sampling saturates quickly. The best score among multiple rollouts is close to the average,
so the behavior itself needs correction.
\emph{Behavior calibration} uses a broad behavior-fix mixture for omitted rubric points, following the
ClinAlign rubric-and-principle synthesis pattern~\citep{clinalign}. It raises full-set scores to
0.601 and 0.332, but the degraded subset falls by 0.252 on Professional and 0.508 on Hard relative
to the base model, so the round is reverted despite its aggregate gains. \emph{Behavior calibration, cleaned}
keeps the same target while removing chart-template boilerplate, adding cleaner principle-rubric
examples, and shortening training; it reaches 0.634 and 0.372. After cross-benchmark consolidation,
Cura 1T reaches the strongest Professional score at 0.662 while remaining close to the
HealthBench-specific round on Hard at 0.368.

\begin{table}[h]
  \caption{HealthBench rubric scores at $T{=}1.0$. Bold marks the best score and underline the
  second-best distinct score in each column.}
  \label{tab:healthbench}
  \centering
  \footnotesize
  \renewcommand{\arraystretch}{1.1}
  \setlength{\tabcolsep}{4pt}
  \begin{tabularx}{0.86\linewidth}{@{}l>{\raggedright\arraybackslash}Xrrl@{}}
    \toprule
    \thd{Model} & \thd{Intervention} & \thd{Professional} & \thd{Hard} & \thd{Decision} \\
    \midrule
    Kimi-K2.6 & $-$ & 0.503 & 0.222 & $-$ \\
    Round 1 & Behavior calibration & 0.601 & 0.332 & revert \\
    Round 2 & Behavior calibration, cleaned & \underline{0.634} & \textbf{0.372} & keep \\
    Cura 1T & Consolidated data mixture & \textbf{0.662} & \underline{0.368} & release \\
    \bottomrule
  \end{tabularx}
\end{table}

\subsection{MedXpertQA}
\label{sec:mxq}

MedXpertQA tests expert medical reasoning across text and multimodal questions. The base model shows
both recoverable reasoning gaps and missing clinical knowledge. Reasoning-only correction helps some
local reasoning patterns but reduces overall accuracy. \emph{Reasoning correction} and \emph{Reasoning correction, extended}
target the diagnosed reasoning gaps with newly synthesized examples, but lack sufficient factual coverage and are reverted, trailing the base model.
\emph{Knowledge injection + retention} adds closed-book clinical knowledge while mixing in examples
the model already answers correctly, raising overall pass@1 to 0.603, and \emph{Data mixture curation}
further balances the retained examples to reach 0.636. \emph{Data mixture curation, capped} is reverted
because overlong traces cause non-termination. The consolidated model reaches 0.655 overall,
improving over the base and Claude Opus 4.8 while remaining second to GPT-5.5.

\begin{table}[h]
  \caption{MedXpertQA split and overall pass@1 at $T{=}1.0$. The overall score weights 2,450 text
  questions and 2,000 multimodal questions. Bold marks the best score and underline the second-best
  distinct score in each column.}
  \label{tab:mxq-detail}
  \centering
  \footnotesize
  \renewcommand{\arraystretch}{1.1}
  \setlength{\tabcolsep}{3pt}
  \begin{tabular}{@{}llrrrl@{}}
    \toprule
    \thd{Model} & \thd{Intervention} & \thd{Text} & \thd{Multimodal} & \thd{Overall} & \thd{Decision} \\
    \midrule
    Kimi-K2.6 & $-$ & 0.484 & 0.672 & 0.569 & $-$ \\
    Round 1 & Reasoning correction & 0.447 & 0.656 & 0.541 & revert \\
    Round 2 & Reasoning correction, extended & 0.454 & 0.657 & 0.545 & revert \\
    Round 3 & Knowledge injection + retention & 0.521 & 0.703 & 0.603 & keep \\
    Round 4 & Data mixture curation & 0.560 & \underline{0.728} & 0.636 & keep \\
    Round 5 & Data mixture curation, capped & $-$ & $-$ & 0.440 & revert \\
    Cura 1T & Consolidated data mixture & \textbf{0.600} & 0.722 & \underline{0.655} & release \\
    \midrule
    Claude Opus 4.8 & $-$ & 0.562 & 0.710 & 0.628 & $-$ \\
    GPT-5.5 & $-$ & \underline{0.596} & \textbf{0.771} & \textbf{0.675} & $-$ \\
    \bottomrule
  \end{tabular}
\end{table}

\subsection{AgentClinic}
\label{sec:agentclinic}

AgentClinic evaluates whether a model can gather evidence and diagnose through a simulated
multi-turn clinical encounter. We adapt the harness into a tool-native format with explicit
structured tools for clinical interaction, so the model acts through structured calls rather than a
text-only dialogue protocol; all rows in \Cref{tab:agentclinic-detail} use this harness at $T{=}1.0$
on MedQA, MedQA-Ext, NEJM, and NEJM-Ext. The main failure mode is premature diagnosis before
completing the clinical workup, especially on NEJM-style subsets. \emph{Single-turn reasoning} is
reverted because isolated answer rationales do not preserve clinical conduct. \emph{Behavior
calibration + retention} instead teaches the model to elicit evidence before committing to a
diagnosis, while mixing new interactive workup examples with retention examples covering behaviors the model already handles; it improves overall pass@1 from 0.754 to 0.807, with the largest gains on NEJM
and NEJM-Ext. Cura 1T preserves most of this gain after consolidation, reaching 0.796.

\begin{table}[h]
  \caption{AgentClinic pass@1 by subset under the tool-native protocol. Bold marks the best score
  and underline the second-best distinct score in each column.}
  \label{tab:agentclinic-detail}
  \centering
  \footnotesize
  \renewcommand{\arraystretch}{1.1}
  \setlength{\tabcolsep}{2.2pt}
  \begin{tabular}{@{}llrrrrrl@{}}
    \toprule
    \thd{Model} & \thd{Intervention} & \thd{MedQA} & \thd{\makecell{MedQA\\Ext}} & \thd{NEJM} & \thd{\makecell{NEJM\\Ext}} & \thd{Overall} & \thd{Decision} \\
    \midrule
    Kimi-K2.6 & $-$ & \underline{0.869} & 0.827 & 0.400 & 0.567 & 0.754 & $-$ \\
    Round 2 & Behavior calibration + retention & 0.841 & 0.841 & \textbf{0.800} & \textbf{0.717} & \textbf{0.807} & keep \\
    Cura 1T & Consolidated data mixture & \textbf{0.879} & \underline{0.850} & \textbf{0.800} & \underline{0.625} & \underline{0.796} & release \\
    \midrule
    Claude Opus 4.8 & $-$ & 0.841 & \textbf{0.874} & \textbf{0.800} & 0.608 & 0.794 & $-$ \\
    GPT-5.5 & $-$ & 0.832 & 0.808 & \underline{0.467} & 0.358 & 0.684 & $-$ \\
    \bottomrule
  \end{tabular}
\end{table}

\subsection{Improvement across benchmarks}

These round-by-round results are development measurements: healthcare benchmark scores guide the keep-or-revise
decisions, while the released model is evaluated after consolidation.

\begin{figure}[h]
  \centering
  \includegraphics[width=0.86\linewidth]{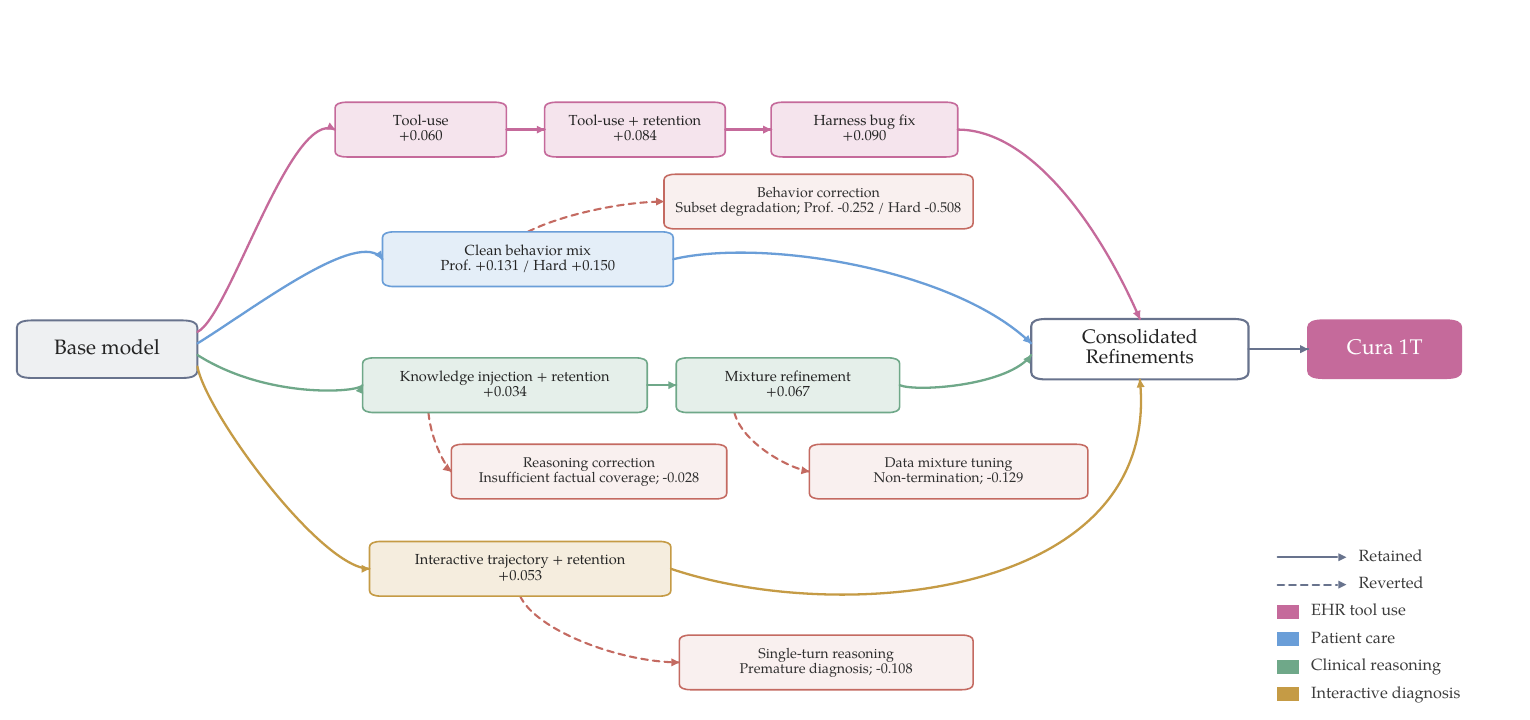}
  \caption{Improvement map from the base model to Cura 1T. Values are changes from benchmark-specific
  bases; solid and dashed red arrows mark retained and reverted interventions.}
  \label{fig:evolution-across-benchmarks}
\end{figure}

\Cref{fig:evolution-across-benchmarks} places the four development paths on a common axis. Two
patterns recur. The intervention that helps is capability-specific, and reverted rounds are a regular
part of the record. In three of the four paths, the first attempt at a capability was reverted before
a working refinement was found.

\subsection{Continual Learning: Out-of-domain Evaluation}
\label{sec:ood}

\begin{figure}[h]
  \centering
  \includegraphics[width=0.66\linewidth]{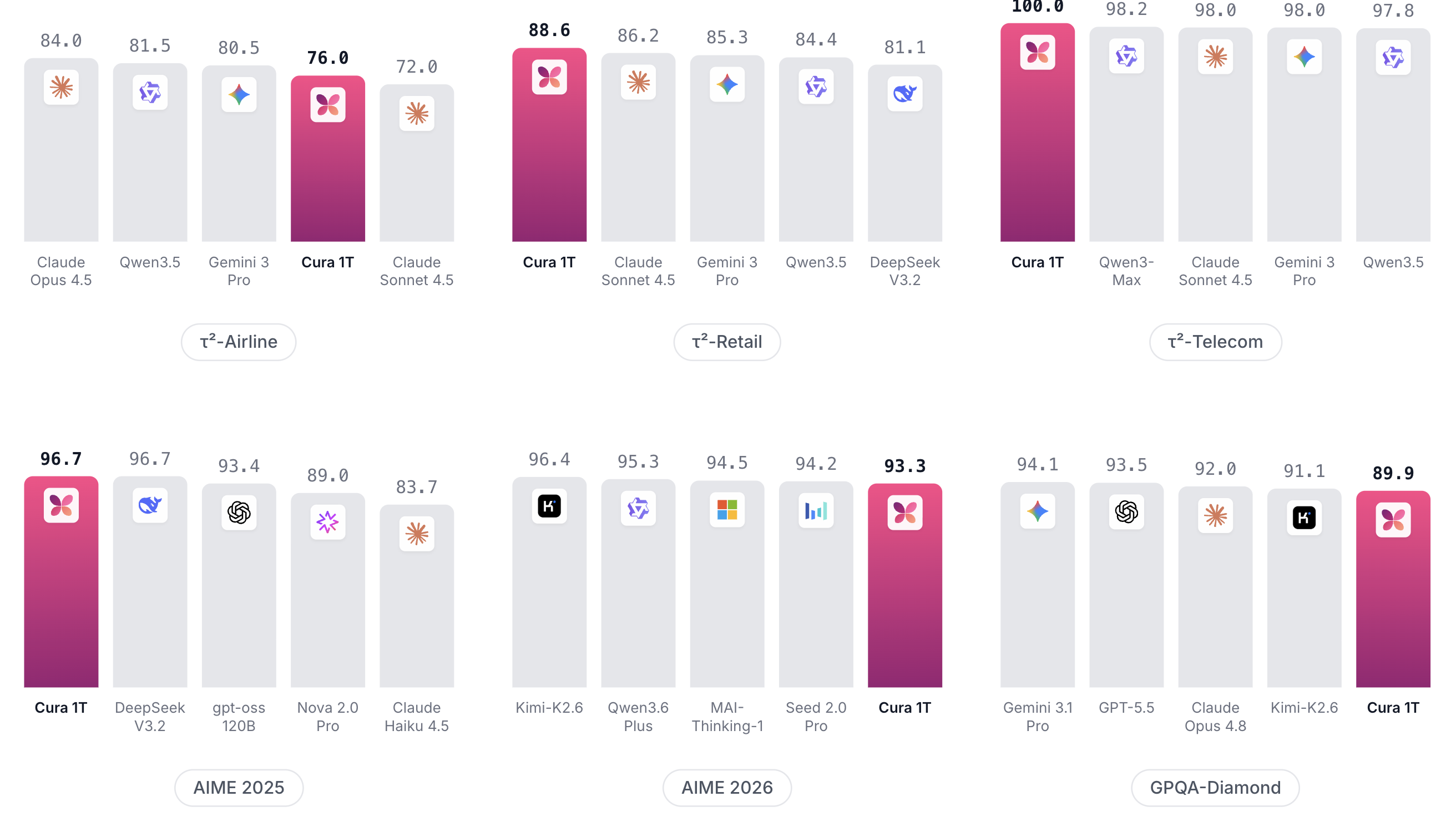}
  \caption{Out-of-domain evaluation results for Cura 1T.}
  \label{fig:ood-comparison}
\end{figure}

We evaluate out-of-domain benchmarks to see whether Cura 1T preserves general capability after
healthcare specialization. These benchmarks were not used to guide the healthcare refinement loops and
provide a separate check of retained general capability. \Cref{fig:ood-comparison} groups them into two families:
\emph{reasoning} covers AIME 2025, AIME 2026, and GPQA-Diamond~\citep{aime,gpqa}, and \emph{agentic}
covers the airline, retail, and telecom domains of $\tau^2$-Bench~\citep{tau2bench}. Cura 1T is on
par with frontier comparators on each reasoning benchmark and on $\tau^2$-airline, and exceeds the
models with publicly reported scores on $\tau^2$-retail and $\tau^2$-telecom.

\section{Discussion}
\label{sec:discussion}

Post-training, especially in specialized domains, was a data-mixture problem more than a hyperparameter-search problem. The
loop treated evaluation failures as evidence for what data was missing, then used SFT to screen
candidate mixtures, RL to improve the policy, and SDFT to consolidate the round. A model may need more rubric-following
behavior for patient care, more factual coverage for expert questions, or more workflow-specific
examples for agentic tool use, and searching the training stack alone would not identify which.

The benchmark traces also show why a single generic medical-data update is insufficient. HealthBench
failures were dominated by omitted rubric items and verbosity-driven grading failures, which
behavior calibration with cleaner principle-rubric data addressed. MedXpertQA mixed
reasoning-pattern failures with missing knowledge, which made reasoning correction alone unstable
until knowledge injection and retention were added. MedAgentBench exposed brittle FHIR writes, while
AgentClinic required interactive trajectories that changed how the model gathered evidence before
answering. Because the failure modes differ, the refinement step categorizes failures before
synthesizing data (\Cref{sec:loop}).

\section{Conclusion}
\label{sec:conclusion}

Cura 1T is a healthcare foundation model built by recursive self-improvement. Starting
from Kimi-K2.6, the loop turns graded benchmark failures into targeted data construction: the agent
plans a round, trains candidate models, evaluates the model, analyzes failed trajectories, and
curates the next mixture, and each accepted round becomes the starting point of the next. The
contribution is a repeatable loop that searches the data mixture.

Across the clinical suite, Cura 1T improves over its base on patient-care, clinical-reasoning, and
healthcare-agentic evaluations, and different settings required different data actions: tool-use
trajectories for MedAgentBench, cleaner behavior data for HealthBench, knowledge injection and
reasoning correction with retention for MedXpertQA, and interactive trajectories for AgentClinic. Out of domain, Cura 1T
matches its comparators on AIME, GPQA-Diamond, and $\tau^2$-airline, and exceeds the models with
published scores on $\tau^2$-retail and $\tau^2$-telecom.

\paragraph{Limitations.} We report the trajectory the loop followed. At trillion-parameter
scale, the cost of isolating the contribution or more ablations is beyond our capabilities. 

\paragraph{Safety and clinical use.} Cura 1T is a research model, not a medical service, and is not a
substitute for a clinician. The benchmarks measure narrow competencies under fixed harnesses, and
strong scores do not establish safety for unsupervised clinical use.

\bibliographystyle{plainnat}
\bibliography{references}

\newpage
\appendix
\section{Recursive Self-Improvement Loop Components}
\label{app:components}

\Cref{tab:components} lists the role and output of each block in \Cref{fig:loop}.

\begin{table}[h]
  \caption{Components of the recursive self-improvement loop in \Cref{fig:loop}. The table summarizes how evaluation records are converted into refined data
  mixtures and how SFT, RL, and SDFT divide the training step.}
  \label{tab:components}
  \centering
  \renewcommand{\arraystretch}{1.15}
  \scriptsize
  \begin{tabularx}{\linewidth}{@{}>{\raggedright\arraybackslash}p{3.55cm}
    >{\raggedright\arraybackslash}X>{\raggedright\arraybackslash}p{3.05cm}@{}}
	    \toprule
    \thd{Component} & \thd{Role in the loop} & \thd{Output} \\
    \midrule
    Plan & Define target behavior, metrics, data recipe, and candidate hyperparameters. & Training specifications \\
    Train & Run SFT to verify the mixture and hyperparameters, RL to improve the policy against reward signals, and SDFT to consolidate the final model. & Candidate model \\
    Evaluate & Run benchmark harnesses and collect graded trajectories and failure summaries. & Failed trajectories and metrics \\
    Refine & Categorize failures, synthesize targeted data, curate the next mixture, validate candidate rows, and suggest next steps. & Validated data refinement \\
    \mbox{Reasoning Correction} & Use a corrected reasoning pattern to synthesize new cases in different clinical contexts. & Reasoning rows \\
    \mbox{Knowledge Injection} & Use verified knowledge to synthesize new questions that require the missing concept. & Source-grounded knowledge rows \\
    \mbox{Behavior Calibration} & Use a rubric or reference behavior to synthesize new cases that exercise the desired response. & Behavior rows \\
    \mbox{Retention Anchor} & Add examples for solved behavior that should be preserved when a correction is added. & Retention rows \\
    Other & Cover task-specific failures outside the reusable reasoning, knowledge, and behavior categories. & Task-specific rows \\
    \mbox{Data Mixture Curation} & Merge, deduplicate, cap, and reweight correction rows with retention anchors. & Candidate training mixture \\
    \mbox{Validation Gates} & Review newly synthesized examples before they enter the training mixture. & Validated training mixture \\
    Human review & Approve the plan, then inspect the validated refinement and candidate result before the next round. & Keep, revert, refine, or deploy decision \\
    \bottomrule
  \end{tabularx}
\end{table}

\section{Training Hyperparameters}
\label{app:training-config}

\Cref{tab:training-config} reports the final training hyperparameters of Cura 1T.
Hyperparameters for these screening runs vary by capability; the table reports the final
consolidation rather than capability-specific screening runs or mixture proportions.

\begin{table}[H]
  \caption{Training hyperparameters for Cura 1T.}
  \label{tab:training-config}
  \centering
  \footnotesize
  \renewcommand{\arraystretch}{1.15}
  \setlength{\tabcolsep}{5pt}
  \begin{tabular}{@{}ll@{}}
    \toprule
    \thd{Configuration} & \thd{Cura 1T} \\
    \midrule
    Algorithm & SFT $\rightarrow$ RL $\rightarrow$ SDFT \\
    Base model & Kimi-K2.6 \\
    Learning rate & $3\times10^{-4}$, linear schedule \\
    Batch size & 128 \\
    Sequence limit & 256K tokens \\
    Training length & 6 epochs with early stopping \\
    \bottomrule
  \end{tabular}
\end{table}

\section{Harness Implementations}
\label{app:harness}

\paragraph{AgentClinic.}
We replace the original free-text control protocol with two native tools. The doctor calls
\texttt{order\_test(test\_name)} to obtain an objective finding and
\texttt{submit\_diagnosis(diagnosis)} to terminate the encounter. History taking remains ordinary
assistant dialogue, which is passed to the patient simulator and returned as a user turn. A test
result can enter the trajectory only through an executed \texttt{order\_test} call, removing the
free-text router's ambiguity between a requested and a reported finding. NEJM images are attached
once in the opening turn, while MedQA encounters remain text-only. The upstream diagnosis comparator
grades the terminal diagnosis, and all models in \Cref{tab:agentclinic-detail} use this interface.

\paragraph{MedAgentBench.}
We expose three renderer-native operations---FHIR read, FHIR write, and final submission---through
\texttt{fhir\_get}, \texttt{fhir\_post}, and \texttt{finish}. Reads are executed against the live FHIR
server. Writes are recorded but do not mutate the server; the grader instead inspects the URL and
resource payload in the tool-call history. The final-submission tool stores the ordered answer values
and terminates the episode. The environment permits up to eight turns, preserves parse failures for
recovery, and scores the completed trajectory with the benchmark's task-specific FHIR graders.

\section{Case Studies}
\label{app:case-studies}

The following selected evaluation traces illustrate changes in model behavior across development rounds.
The cases are held out from training and are reported as qualitative companions to the aggregate results.

\subsection{MedAgentBench: Referral ServiceRequest}

\paragraph{Task and criterion.}
Task \texttt{task8\_1} asks for an orthopedic surgery referral ServiceRequest for synthetic patient
S2016972 at timestamp 2023-11-13T10:15:00\allowbreak+00:00. The grader requires SNOMED
306181000000106, \texttt{active} status, \texttt{order} intent, \texttt{stat} priority, the exact
subject reference, and an SBAR note.

\begin{table}[H]
  \centering
  \footnotesize
  \renewcommand{\arraystretch}{1.15}
  \begin{tabularx}{\linewidth}{@{}>{\raggedright\arraybackslash}p{3.0cm}>{\raggedright\arraybackslash}Xl@{}}
    \toprule
    \thd{Stage} & \thd{Response excerpt} & \thd{Result} \\
    \midrule
    Base & \texttt{fhir\_post(ServiceRequest, priority=\textcolor{curaRed}{routine}, code=306181000000106, display=...)} & Reward 0 \\
    Round 1 (Behavior calibration) & \texttt{fhir\_post(ServiceRequest, priority=stat, code=306181000000106)} & Reward 1 \\
    Round 2 (Behavior calibration + retention) & \texttt{priority=stat}; exact SNOMED coding, subject, timestamp, and SBAR note are preserved. & Reward 1 \\
    Round 3 (Harness bug fix) & Under the corrected harness: \texttt{priority=\textcolor{curaGreen}{stat}} with exact coding and \texttt{finish([])}. & Reward 1 \\
    \bottomrule
  \end{tabularx}
\end{table}

\paragraph{Change.}
The base model understands the referral intent but misses the exact FHIR contract. Later responses
satisfy the schema and content constraints, and the harness bug fix confirms the correction under
the intended evaluation protocol.

\subsection{HealthBench Professional: EROSTIN 10MG}

\paragraph{Task and criterion.}
The prompt asks, ``what medication is EROSTIN 10MG''. The criterion requires identifying ebastine as
the active ingredient and avoiding confusion with estradiol or Elestrin.

\begin{table}[H]
  \centering
  \footnotesize
  \renewcommand{\arraystretch}{1.15}
  \begin{tabularx}{\linewidth}{@{}>{\raggedright\arraybackslash}p{3.0cm}>{\raggedright\arraybackslash}Xl@{}}
    \toprule
    \thd{Stage} & \thd{Response excerpt} & \thd{Result} \\
    \midrule
    Base & ``I \textcolor{curaRed}{cannot definitively identify} what drug `Erostin 10 mg' is based on the brand name alone.'' & Reward 0 \\
    Behavior calibration & ``EROSTIN 10 mg is an antihistamine medication containing ebastine.'' The task is correct, although the round is reverted for subset-level degradation. & Reward 1 \\
    Retention check & ``EROSTIN 10 mg is an antihistamine medication that contains ebastine as its active component.'' & Reward 1 \\
    Cura 1T & ``EROSTIN 10 mg is an antihistamine medication that contains \textcolor{curaGreen}{ebastine} as its active ingredient.'' & Reward 1 \\
    \bottomrule
  \end{tabularx}
\end{table}

\paragraph{Change.}
This is a drug-identification failure rather than a reasoning failure. Behavior calibration answers
this example locally but is not retained globally; the intermediate retention check and Cura
1T preserve the corrected active ingredient. A matched trace for the cleaned round is not available
for this example, so it is not shown.

\subsection{HealthBench Professional: Sciatica Coding Regression}

\paragraph{Task and criterion.}
The prompt asks for the current ICD-10-CM coding for sciatica. A correct answer must distinguish the
billable laterality codes: \texttt{M54.30} for unspecified side, \texttt{M54.31} for right, and
\texttt{M54.32} for left. The parent category \texttt{M54.3} is not the complete unspecified-side
code.

\begin{table}[H]
  \centering
  \footnotesize
  \renewcommand{\arraystretch}{1.15}
  \begin{tabularx}{\linewidth}{@{}>{\raggedright\arraybackslash}p{3.0cm}>{\raggedright\arraybackslash}Xl@{}}
    \toprule
    \thd{Stage} & \thd{Response excerpt} & \thd{Result} \\
    \midrule
    Base & ``\texttt{M54.30}---sciatica, unspecified side; \texttt{M54.31}---right; \texttt{M54.32}---left.'' & Reward 1.000 \\
    Behavior calibration & ``\texttt{M54.3}---sciatica, unspecified side,'' while retaining \texttt{M54.31} and \texttt{M54.32}. & Reward 0.000 \\
    Cura 1T & Restores \texttt{M54.30}, \texttt{M54.31}, and \texttt{M54.32}, and distinguishes ICD-10-CM from the WHO category. & Reward 0.973 \\
    \bottomrule
  \end{tabularx}
\end{table}

\paragraph{Change.}
Behavior calibration preserves the correct code family but drops the final digit for unspecified
laterality, turning a correct base response into an incomplete coding recommendation. Cura 1T
restores the billable code and the laterality distinction. This matched trace illustrates the
subset-level degradation that caused Round 1 to be reverted despite its higher overall score.

\subsection{MedXpertQA Text-197: DCIS Radiotherapy Benefit}

\paragraph{Task and criterion.}
The question asks which ductal carcinoma in situ patient gains the greatest local-control benefit
from radiotherapy. The gold answer is D: a 0.9 cm grade-3 DCIS after lumpectomy with a negative
margin.

\begin{table}[H]
  \centering
  \footnotesize
  \renewcommand{\arraystretch}{1.15}
  \begin{tabularx}{\linewidth}{@{}>{\raggedright\arraybackslash}p{3.0cm}>{\raggedright\arraybackslash}Xl@{}}
    \toprule
    \thd{Stage} & \thd{Response excerpt} & \thd{Result} \\
    \midrule
    Base & ``A positive surgical margin signifies likely residual disease and the highest baseline recurrence risk~\ldots{} \(\boxed{\textcolor{curaRed}{\mathrm{F}}}\).'' & Reward 0 \\
    Round 1 (Reasoning correction) & ``Option F combines~\ldots{} grade 3 histology and a positive margin. \(\boxed{\mathrm{F}}\).'' & Reward 0 \\
    Round 2 (extended) & ``The highest-risk breast-conservation patient~\ldots{} \(\boxed{\mathrm{F}}\).'' & Reward 0 \\
    Round 3 (Knowledge injection + retention) & ``Positive-margin cases~\ldots{} are not rescued by RT alone~\ldots{} Option D~\ldots{} \(\boxed{\mathrm{D}}\).'' & Reward 1 \\
    Round 4 (Data mixture curation) & ``Positive-margin cases~\ldots{} are not rescued by RT alone~\ldots{} Option D~\ldots{} \(\boxed{\mathrm{D}}\).'' & Reward 1 \\
    Cura 1T & ``Positive margins~\ldots{} require further surgery~\ldots{} Option \textcolor{curaGreen}{D}~\ldots{} \(\boxed{\mathrm{D}}\).'' & Reward 1 \\
    \bottomrule
  \end{tabularx}
\end{table}

\paragraph{Change.}
The missing piece is a clinical constraint: positive margins change the management problem instead
of simply increasing radiotherapy benefit. The retained rounds correct that constraint and preserve
the answer through consolidation.

\subsection{AgentClinic NEJM-Ext 21: Desquamative Interstitial Pneumonia}

\paragraph{Task and criterion.}
The case describes a 34-year-old woman with tobacco use, dyspnea and dry cough, diffuse ground-glass
opacities with peripheral consolidation, nondiagnostic bronchoalveolar lavage, and biopsy showing
pigment-laden macrophages. The gold diagnosis is desquamative interstitial pneumonia.

\begin{table}[H]
  \centering
  \footnotesize
  \renewcommand{\arraystretch}{1.15}
  \begin{tabularx}{\linewidth}{@{}>{\raggedright\arraybackslash}p{3.0cm}>{\raggedright\arraybackslash}Xl@{}}
    \toprule
    \thd{Stage} & \thd{Response excerpt} & \thd{Result} \\
    \midrule
    Base & ``The most likely diagnosis is \textcolor{curaRed}{Chronic Eosinophilic Pneumonia}.'' & Reward 0 \\
    Reverted diagnostic attempt & ``DIAGNOSIS READY: Pulmonary alveolar proteinosis,'' despite normal anti-GM-CSF and nondiagnostic BAL/PAS results. & Reward 0 \\
    Behavior calibration + retention & ``The diffuse ground-glass opacities~\ldots{} and macrophages in the airspaces~\ldots{} [are] classic for Desquamative Interstitial Pneumonia.'' & Reward 1 \\
    Cura 1T & Requests biopsy review, receives ``Extensive alveolar filling with pigment-laden macrophages,'' and submits \textcolor{curaGreen}{desquamative interstitial pneumonia}. & Reward 1 \\
    \bottomrule
  \end{tabularx}
\end{table}

\paragraph{Change.}
This multi-turn case depends on gathering and using discriminating evidence before diagnosing. The
retained intervention improves the workup behavior, and Cura 1T preserves that evidence-led
diagnosis.

\end{document}